\documentclass[twoside,11pt]{article}

\usepackage{jmlr2e}
\usepackage{subcaption}

\firstpageno{1}

\begin{document}

\title{Predicting Inpatient Discharge Prioritization With Electronic Health Records}

\author{\name Anand Avati \email avati@cs.stanford.edu
    \AND
    \name Stephen Pfohl \email spfohl@stanford.edu
    \AND
    \name Chris Lin \email clin17@stanford.edu
    \AND
    \name Thao Nguyen \email thaonguyen@cs.stanford.edu
    \AND
    \name Meng Zhang \email meng.zhang@stanford.edu
    \AND
    \name Philip Hwang \email phwang20@stanford.edu
    \AND
    \name Jessica Wetstone \email wetstone@stanford.edu
    \AND
    \name Kenneth Jung \email kjung@stanford.edu
    \AND
    \name Andrew Ng \email ang@cs.stanford.edu
    \AND
    \name Nigam H. Shah \email nigam@stanford.edu}

\maketitle

\begin{abstract}

Identifying patients who will be discharged within 24 hours can improve hospital resource management and quality of care. We studied this problem using eight years of Electronic Health Records (EHR) data from Stanford Hospital. We fit models to predict 24 hour discharge across the entire inpatient population.  The best performing models achieved an area under the receiver-operator characteristic curve (AUROC) of 0.85 and an AUPRC of 0.53 on a held out test set.  This model was also well calibrated.  Finally, we analyzed the utility of this model in a decision theoretic framework to identify regions of ROC space in which using the model increases expected utility compared to the trivial always negative or always positive classifiers.

\end{abstract}

\section{Introduction}

In 2016 the average daily cost of an inpatient stay was \$3,421 in California, and has continued to rise since then \citep{InpatientCost}. As hospitals try to lower costs and improve quality care, operational goals such as the timely discharge of patients who are ready to leave the hospital have become increasingly important.  Timely discharge provides many benefits to patients and health care systems by freeing scarce beds for other patients and reducing exposure to iatrogenic conditions \citep{hospitalAcquired}, in addition to reducing costs per patient, improving post-discharge care \citep{postdischarge}, and lowering risk of readmission \citep{losreadmission}. In this work, we present preliminary work on predicting which patients who will be discharged in the next 24 hours. Accurate identification of such patients can help hospitals reduce unnecessary delays that incur additional days in the hospital for discharge ready patients by prioritizing these patients for critical services that must be done prior to discharge (e.g., imaging, lab tests, arranging for transportation) and \emph{load balancing} so that each discharge ready patient is under the care of different clinical staff.  

We approach this task by formulating a supervised learning problem, taking advantage of the abundance of patient data available in Electronic Health Records (EHR).  Such data have been successfully applied to a variety of clinical tasks, such as medical imaging diagnosis \citep{gulshan2016development}, mining medical notes \citep{nguyen2017mathtt}, and disease onset prediction \citep{liu2018deepehr}. We compare the performance of various machine learning models on the task of predicting 24 hour discharge.  Our best performing model, a gradient boosted tree model, achieves good discrimination and calibration on held out test data.  We also perform a preliminary analysis of the marginal utility gains possible relative to trivial classifiers.  We find that there is some evidence that these models, with appropriately designed down stream interventions, may offer some benefit.  

\section{Related Work}
Widespread adoption of EHR systems following the passage of the Affordable Care Act \citep{ehradoption} has fueled increasing interest in the secondary use of data collected during routine clinical care as input to machine learning algorithms to solve clinical problems. Prior work has focused on predicting diverse clinical outcomes such as prolonged length of stay, unplanned readmissions, inpatient mortality and diagnoses \citep{GoogleEHR2018, multilearn, multisubgroups, multidisease}.  There is also substantial prior work focusing on forecasting length of stay and discharge, but this work has typically been limited to specific patient populations. \citep{chaou2017predicting} modelled Emergency Department LOS using accelerated failure time model and a relatively specific set of features, including triage level index as assigned by a specialized nurse, while \citep{acslos} focused on predicting LOS for acute coronary syndrome patients.  \citep{barnes2015real} sought to produce daily predictions of discharge for a single inpatient unit. Similar attempts were made to study the same problem for cardiac patients in \citep{cardiaclos}. In contrast, our present work encompasses the entire inpatient population. 

\section{Data}
\subsection{Data Source}
We used EHR data from STRIDE (Stanford Translational Research Integrated Database Environment) \citep{lowe2009stride}, a clinical data warehouse containing demographic and clinical information on 3.3 million pediatric and adult patients at Stanford University Medical Center.  

\subsection{Cohort Selection}
EHR data on approximately 1 million adult patients at SHC were extracted from encounters occurring between January 1, 2010 and February 10, 2018. Nested encounters (encounters that happened during other encounters) were removed. Encounters from the same patient that happened within twelve hours were merged into a single encounter. 

To approximate real-world deployment in our model evaluation, patients with inpatient encounters before Jan 1, 2017 were used for training, while those with inpatient encounters on or after Jan 1, 2017 were randomly split 50-50 into validation and test sets. For patients with multiple inpatient visits, only one visit was randomly selected in the validation and test set while all visits were preserved in the training set. To further safeguard the integrity of test data and avoid potential data leakage, we removed all the patients that are included in the testing set from the training set. The prevalence of positive cases - inpatient discharge in the next day - is around 18\% in all split data sets (Table~\ref{tab:data 
split}).

    

\begin{table}[htbp]
  \centering 
  \caption{Statistics of the data split used for modeling} 
  \resizebox{\textwidth}{!}{\begin{tabular}{|l|c|c|c|c|c|}\hline
     & Time Period & Number of visits & Number of patients & Days in hospital \\ \hline
    Training & Jan 1, 2010 - Jan 1, 2017 & 6,997,831 & 83,797 & 7,081,628 \\ \hline
    Validation & Jan 1, 2017 - Feb 10, 2018 & 852,554 & 7,212 & 859,766 \\ \hline
    Testing & Jan 1, 2017 - Feb 10, 2018 & 876,572 & 7,211 & 883,783  \\ \hline
    
  \end{tabular}}
  \label{tab:data split} 
\end{table}

\section{Methods}
We formulated a supervised learning problem in which the task is to map health care data for a given patient collected in the EHR prior to a specific day to the probability of the patient being discharged in the next 24 hours. 

\subsection{Patient Representation}
\subsubsection{Data Elements}
We used diagnosis codes, procedure codes, medication codes, lab codes, and the corresponding lab results (categorized as normal, abnormal, or panic) from encounter records, procedure records, medication orders, and lab result records respectively. Diagnosis and procedure codes were ICD-9 \citep{icd9} and CPT \citep{cpt} codes respectively. Codes with fewer than 100 occurrences in the training set were excluded. Patient age, gender, race, ethnicity, insurance type, and whether the visit is for surgery were also included as background information. 

\subsubsection{Fixed Length Patient Representations}
\begin{figure}[htbp]
  \centering 
  \includegraphics[width=3in]{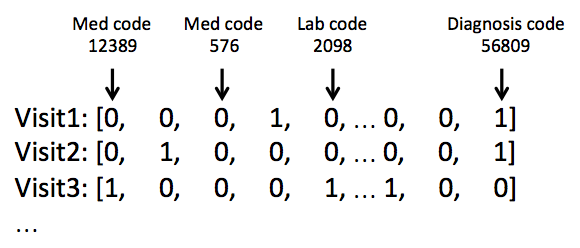} 
  \caption{Schematic view of data inputs (visit events are not necessarily in temporal order)}
  \label{fig:baselinefeatures} 
\end{figure}

Many machine learning models, such as logistic regression and random forests, require a fixed length vector of input data.  For such models we must summarize the longitudinal medical history of each patient into such a fixed length vector. For each day of an inpatient visit, the clinical codes and lab results from the day before were counted and considered recent data, while those that occurred 2 days to 6 months before were aggregated and considered historical data. Historical events (diagnosis codes, procedure codes, medication codes, lab codes, and lab results) were aggregated by summing up occurrences over the entire historical record, discarding their temporal ordering (Fig.~\ref{fig:baselinefeatures}).The two sets of past data, together with the patient demographics, were used as input features for models requiring fixed length representations.

\begin{figure}[htbp]
  \centering 
  \includegraphics[width=6in]{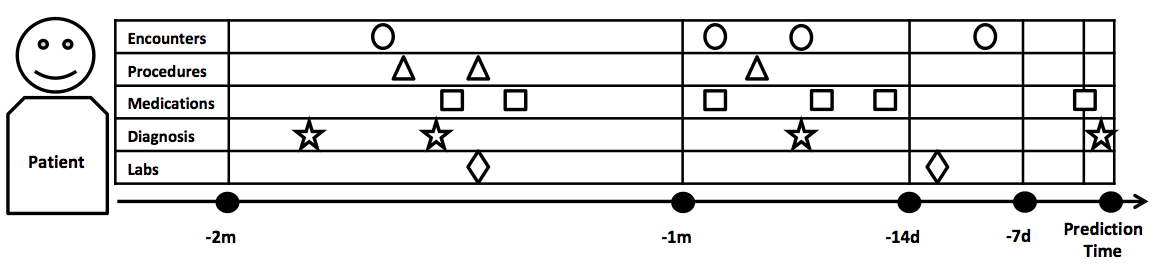} 
  \caption{Schematic view of sequential data inputs, with random time markers chosen to illustrate the history of the patient}
  \label{fig:rnnfeatures} 
\end{figure} 

\subsubsection{Data Inputs for Recurrent Neural Networks}
In contrast, recurrent neural nets are able to handle variable length inputs, and in particular,  variable length longitudinal medical records.  For each patient, the same set of features described above were grouped by day, and fed into RNN as an ordered sequence, representing the timeline of that patient's interaction with the hospital (Fig.~\ref{fig:rnnfeatures}). For more efficient learning updates, patients with similar total length of stay were batched together during training.

\subsection{Models}
\subsubsection{Random Forest}
Random Forests pool the responses of an ensemble of decision tree classifiers, each trained on bootstrapped versions of the original dataset. After hyperparameter-tuning using the validation set, we obtained a Random Forest with 2,000 classification trees. All classification trees were grown until all leaves contained fewer than two samples. Random Forests were constructed using the scikit-learn library (version 0.19.0) in Python (version 3.6.1).

\subsubsection{Gradient Boosting}
Gradient Boosting Machines (GBM) are another form of decision tree ensemble, but it builds the ensemble by iterative functional gradient descent. There are many implementations of gradient boosting, each of which uses various heuristics to both regularize and increase computational efficiency.  In this study, we used two implementations - scikit-learn's standard gradient boosted classifier, and XGBoost.  For the former, we used 500 component trees, a sub-sampling fraction of 0.8, and considered 482 features at each split.  Each classification tree was grown to a maximum depth of 50 or a a minimum of 3 samples per leaf node, and we used a learning rate of 0.1  For XGBoost, we used an ensemble of 2000 classifiers, with a minimum of 2 samples per leaf node and a learning rate of 0.3.  

\subsubsection{Recurrent Neural Nets}
We used a recurrent neural net comprising one recurrent layer of 64 Gated Recurrent Unit (GRU) \citep{gru} hidden units.  Diagnosis codes, procedure codes, medication codes, lab codes, and encounter types were first passed through an embedding layer. We experimented with two different approaches for embedding: (i) embed each feature separately as a vector of size 25, (ii) dividing the features into 2 groups based on "internal" factors (diagnosis, lab) and "external" factors (medication, encounter) affecting the patient's conditions, and embed each group as a vector of size 50. The outputs of the embedding layer at each time step were averaged and then concatenated with demographics and visit features before being fed as inputs into the recurrent layer. We used a fully connected layer before a sigmoid activation function at the output layer.  The model was trained with the Adam optimizer for 20 epochs, a learning rate of 0.003 and a dropout probability of 0.2 across the network. No weight decay was used. We also used auxiliary targets, i.e.,  multitask learning, to improve learning.  At each time step, we updated gradients based on cross entropy loss from one of the three prediction tasks: (i) whether the patient is discharged in the next 24 hours, (ii) whether the patient was inpatient at the existing time step, (iii) whether the patient would be inpatient in the next time step. The last 2 auxiliary tasks used data from all patients instead of just inpatients. They were chosen because it was not difficult to integrate them into the existing data processing and loading pipeline, and we hypothesized that they would help generate more robust representations that correlate with 24-hour discharge.

\begin{figure}
\centering
\begin{minipage}{.5\textwidth}
  \centering
  \includegraphics[width=1\linewidth]{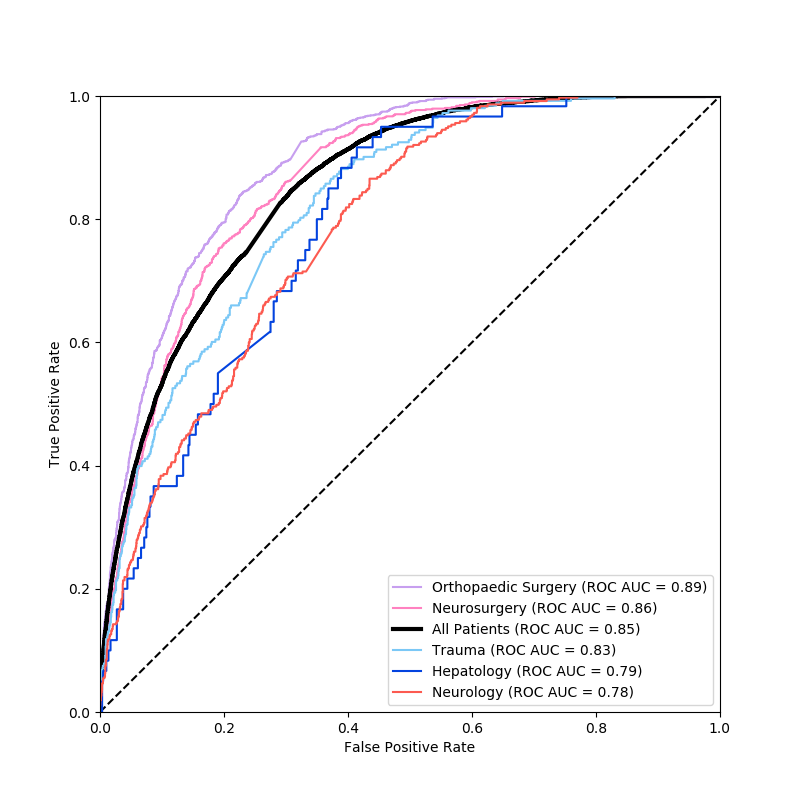}
  \captionof{figure}{ROC Curve of XGBoost model \\
  on test set}
  \label{fig:ROC}
\end{minipage}%
\begin{minipage}{.5\textwidth}
  \centering
  \includegraphics[width=1\linewidth]{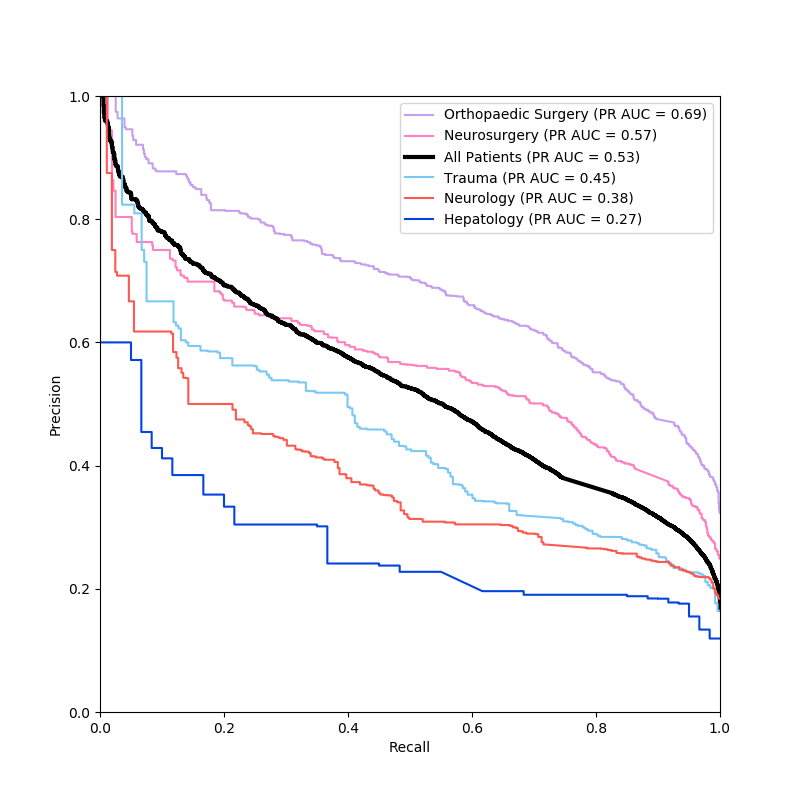}
  \captionof{figure}{Interpolated Precision-Recall Curve of XGBoost model on test set}
  \label{fig:PRC}
\end{minipage}
\end{figure}

\section{Results} 
\subsection{Evaluation Metrics} 
The primary metric used for model selection is Area Under the Receiver Operating 
Characteristic Curve (AUROC), a measure of model's discrimination ability between patients who will be discharged and those who will not, at different classification thresholds. We also compared models by the Area Under the Precision-Recall Curve (AUPRC).  For the best-performing model, we evaluated its calibration by calculating the Brier Score, which measures how close the predicted probabilities are to the true probabilities. This is estimated calculating the concordance between the empirical probability of discharge in bins of predicted probabilities output by the model.

\begin{table}[htbp]
  \centering 
  \caption{Test performance for 24 hr discharge} 
  \begin{tabular}{|l|l|l|l|}\hline
    Model & AUROC & AUPRC & Calibration \\ \hline
    Random Forest & 0.80 & 0.46 & 0.11 \\ \hline
    GBM & 0.84 & 0.48 & 0.11 \\ \hline
    \textbf{XGBoost} & \textbf{0.85} & \textbf{0.53} & \textbf{0.10} \\ \hline
    GRU & 0.82 & 0.50 & 0.11 \\ \hline
  \end{tabular}
  \label{tab:disch model performance}
\end{table}

\begin{figure}[htbp]
  \centering 
  \includegraphics[width=4in]{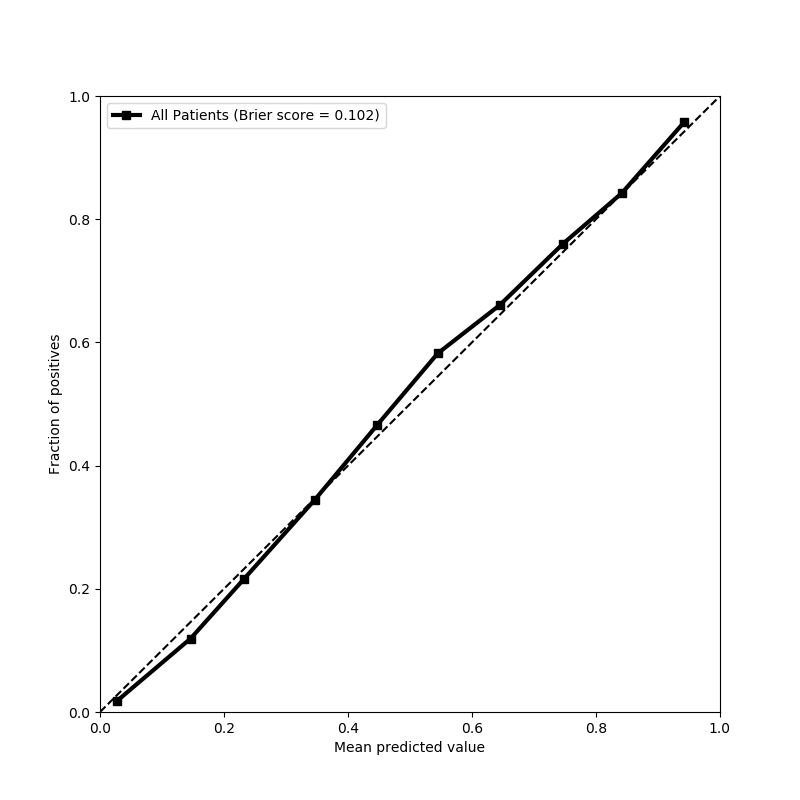} 
  \caption{Calibration plot of XGBoost output probabilities on the test set}
  \label{fig:calibration} 
\end{figure} 
\subsection{Prediction Results on All Patients} 

Results across all models is shown in table \ref{tab:disch model performance}.  The models achieved comparable performance, with XGBoost achieving the best AUROC (0.85 vs 0.84 for the second best model) and AUPRC (0.53 vs 0.50 for the second best model). All models were well calibrated, with Brier scores ranging from 0.11 to 0.102 for XGBoost. Fig.\ref{fig:calibration} shows the calibration plot for the XGBoost model. Given a classification threshold of 0.5, the probability estimates were conservative for all the positive discharge predictions. Thus, the imperfect calibration would not affect the ranking of patient priority and should not pose a concern during clinical deployment.

\subsection{Prediction Results on Service Lines}
We worked with clinicians from SHC to identify specific service lines where discharge protocols are standardized and straightforward, and thus where our model's predictions are most actionable. 
The XGBoost model performance for these service lines is shown in Fig.~\ref{fig:ROC} and Fig.~\ref{fig:PRC}. We observe that the predictions in Orthopaedic Surgery and Neurosurgery have the highest AUROC and AUPRC respectively.  

\subsection{Analysis of Marginal Utility Gains}
\begin{figure}[htbp]
  \centering
  \includegraphics[width=3.8in]{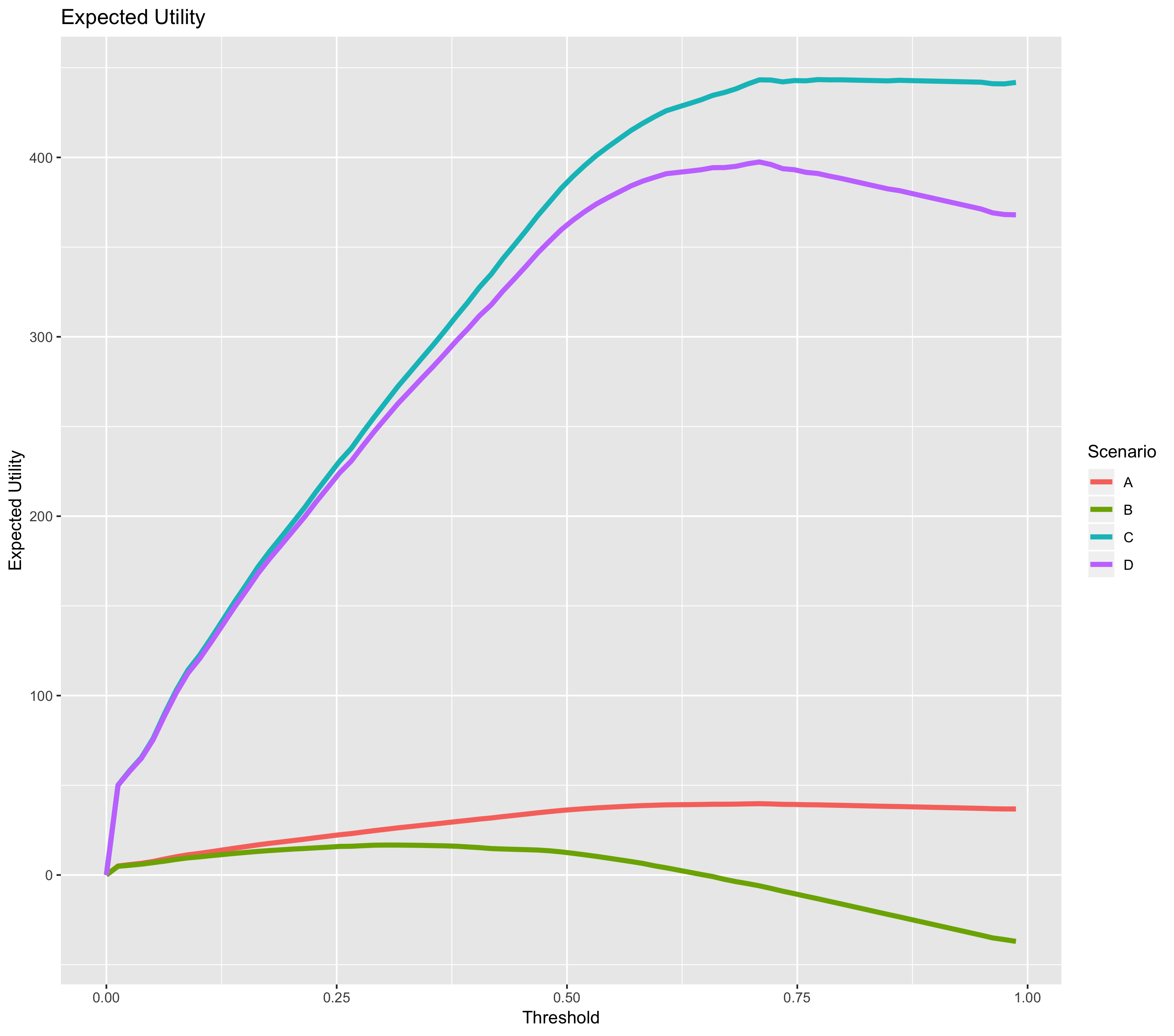}
  \caption{Expected Utility as a function of classifier threshold under four scenarios, characterized by whether the cost of a false positive is high or low, and whether the benefit of a true positive is high or low.  Respectively, scenario A is (low, low), B is (high, low), C is (low, high), and D is (high, high).}
  \label{fig:expected_utility}
\end{figure}

In this section, we estimate whether our best model provides any benefit in terms of marginal expected utility. 
We fix the prevalence of 24 hour discharge as 18\% and make the following assumptions about the utility of false positives and true positives.  For false positives, we assume that a reasonable range of the cost of false positives relative to true negatives is -\$10 to -\$100.  For true positives, we assume that the benefit of true positives relative to false negatives is in the range \$250 to \$2500, with the upper bound set to approximately the average cost of an inpatient day in California.  Given these assumptions, we can calculate the expected utility of our best model under four scenarios, one for each combination of utility differences.  Fig.~\ref{fig:expected_utility} shows expected utility as a function of decision threshold under these four scenarios.  Under the most optimistic scenarios C and D, in which benefit of a true positive is large (\$2500), the expected utility is positive and large at all thresholds.  Even under the most pessimistic scenario, B, which corresponds to the case where the cost of a false positive is high (-\$100) and the benefit of a true positive is low (\$250), this analysis indicates that we can set our decision threshold to a value that achieves a positive marginal utility.  

\section{Discussion} 
In this work, we compare machine learning models for predicting inpatients who will be discharged within 24 hours.  Such models may be used to increase the rate of timely discharge by prioritizing them for discharge related services and other interventions that help ensure timely discharge. The best performing model, XGBoost, achieved an AUROC of 0.85 and was well calibrated. A preliminary analysis of the utility gains possible with this model suggests that it could, with appropriate downstream actions, lead to a gain in expected utility.  Note that our models were developed and evaluated on data from a single medical center, and we thus cannot claim that such results are possible elsewhere.  We also acknowledge that our cost benefit analysis relies on possibly unrealistic assumptions.  These limitations notwithstanding, we believe these results are encouraging and suggest that predictive models for 24 hour discharge have potential to provide benefit to the health care system.  In future work, we will elicit estimates of the relevant utilities from domain experts, and perform simulations in order to obtain better estimates of the utility of our models and suggest optimal decision thresholds for our models prior to a prospective trial at Stanford Hospital.  





\bibliographystyle{natbib}
\bibliography{main}

\begin{thebibliography}{20}
\providecommand{\natexlab}[1]{#1}
\providecommand{\url}[1]{\texttt{#1}}
\expandafter\ifx\csname urlstyle\endcsname\relax
  \providecommand{\doi}[1]{doi: #1}\else
  \providecommand{\doi}{doi: \begingroup \urlstyle{rm}\Url}\fi

\bibitem[Adler-Milstein et~al.(2015)Adler-Milstein, DesRoches, Kralovec,
  Foster, Worzala, Charles, Searcy, and Jha]{ehradoption}
Julia Adler-Milstein, Catherine~M DesRoches, Peter Kralovec, Gregory Foster,
  Chantal Worzala, Dustin Charles, Talisha Searcy, and Ashish~K Jha.
\newblock Electronic health record adoption in us hospitals: progress
  continues, but challenges persist.
\newblock \emph{Health affairs}, 34\penalty0 (12):\penalty0 2174--2180, 2015.

\bibitem[AHA(2016)]{InpatientCost}
AHA.
\newblock Hospital adjusted expenses per inpatient day, 2016.
\newblock URL
  \url{https://www.kff.org/health-costs/state-indicator/expenses-per-inpatient-day/}.

\bibitem[Association(2007)]{cpt}
American~Medical Association.
\newblock \emph{Current procedural terminology: CPT}.
\newblock American Medical Association, 2007.

\bibitem[Barnes et~al.(2015)Barnes, Hamrock, Toerper, Siddiqui, and
  Levin]{barnes2015real}
Sean Barnes, Eric Hamrock, Matthew Toerper, Sauleh Siddiqui, and Scott Levin.
\newblock Real-time prediction of inpatient length of stay for discharge
  prioritization.
\newblock \emph{Journal of the American Medical Informatics Association},
  23\penalty0 (e1):\penalty0 e2--e10, 2015.

\bibitem[Chaou et~al.(2017)Chaou, Chen, Chang, Tang, Pan, Yen, and
  Chiu]{chaou2017predicting}
Chung-Hsien Chaou, Hsiu-Hsi Chen, Shu-Hui Chang, Petrus Tang, Shin-Liang Pan,
  Amy Ming-Fang Yen, and Te-Fa Chiu.
\newblock Predicting length of stay among patients discharged from the
  emergency department—using an accelerated failure time model.
\newblock \emph{PloS one}, 12\penalty0 (1):\penalty0 e0165756, 2017.

\bibitem[Cho et~al.(2014)Cho, Van~Merri{\"e}nboer, Gulcehre, Bahdanau,
  Bougares, Schwenk, and Bengio]{gru}
Kyunghyun Cho, Bart Van~Merri{\"e}nboer, Caglar Gulcehre, Dzmitry Bahdanau,
  Fethi Bougares, Holger Schwenk, and Yoshua Bengio.
\newblock Learning phrase representations using rnn encoder-decoder for
  statistical machine translation.
\newblock \emph{arXiv preprint arXiv:1406.1078}, 2014.

\bibitem[Colwell(2014)]{postdischarge}
Janet Colwell.
\newblock Length of stay: Timing it right. strategies for achieving efficient,
  high-quality care.
\newblock \emph{ACP Hospitalist}, 2014.
\newblock URL
  \url{https://www.theguardian.com/world/2017/mar/12/netherlands-will-pay-the-price-for-blocking-turkish-visit-erdogan}.

\bibitem[for Disease~Control et~al.(2013)for Disease~Control, Prevention,
  et~al.]{icd9}
Centers for Disease~Control, Prevention, et~al.
\newblock International classification of diseases, ninth revision, clinical
  modification (icd-9-cm).
\newblock \emph{Atlanta, Georgia, USA. Available on: http://www. cdc.
  gov/nchs/icd/icd9cm. htm}, 2013.

\bibitem[Gulshan et~al.(2016)Gulshan, Peng, Coram, Stumpe, Wu, Narayanaswamy,
  Venugopalan, Widner, Madams, Cuadros, et~al.]{gulshan2016development}
Varun Gulshan, Lily Peng, Marc Coram, Martin~C Stumpe, Derek Wu, Arunachalam
  Narayanaswamy, Subhashini Venugopalan, Kasumi Widner, Tom Madams, Jorge
  Cuadros, et~al.
\newblock Development and validation of a deep learning algorithm for detection
  of diabetic retinopathy in retinal fundus photographs.
\newblock \emph{Jama}, 316\penalty0 (22):\penalty0 2402--2410, 2016.

\bibitem[Hachesu et~al.(2013)Hachesu, Ahmadi, Alizadeh, and
  Sadoughi]{cardiaclos}
Peyman~Rezaei Hachesu, Maryam Ahmadi, Somayyeh Alizadeh, and Farahnaz Sadoughi.
\newblock Use of data mining techniques to determine and predict length of stay
  of cardiac patients.
\newblock \emph{Healthcare informatics research}, 19\penalty0 (2):\penalty0
  121--129, 2013.

\bibitem[Harutyunyan et~al.(2017)Harutyunyan, Khachatrian, Kale, and
  Galstyan]{multilearn}
Hrayr Harutyunyan, Hrant Khachatrian, David~C Kale, and Aram Galstyan.
\newblock Multitask learning and benchmarking with clinical time series data.
\newblock \emph{arXiv preprint arXiv:1703.07771}, 2017.

\bibitem[J.~Graham~Atkinson(2014)]{hospitalAcquired}
D.Phil. J.~Graham~Atkinson.
\newblock The relationship between length of stay and the probability of
  incurring a hospital complication: a two-way interaction.
\newblock 2014.

\bibitem[Kaboli et~al.(2012)Kaboli, Go, Hockenberry, Glasgow, Johnson,
  Rosenthal, Jones, and Vaughan-Sarrazin]{losreadmission}
Peter~J Kaboli, Jorge~T Go, Jason Hockenberry, Justin~M Glasgow, Skyler~R
  Johnson, Gary~E Rosenthal, Michael~P Jones, and Mary Vaughan-Sarrazin.
\newblock Associations between reduced hospital length of stay and 30-day
  readmission rate and mortality: 14-year experience in 129 veterans affairs
  hospitals.
\newblock \emph{Annals of internal medicine}, 157\penalty0 (12):\penalty0
  837--845, 2012.

\bibitem[{Liu} et~al.(2018){Liu}, {Zhang}, and {Razavian}]{liu2018deepehr}
J.~{Liu}, Z.~{Zhang}, and N.~{Razavian}.
\newblock {Deep EHR: Chronic Disease Prediction Using Medical Notes}.
\newblock \emph{ArXiv e-prints}, August 2018.

\bibitem[Lowe et~al.(2009)Lowe, Ferris, Hernandez, and Weber]{lowe2009stride}
Henry~J Lowe, Todd~A Ferris, Penni~M Hernandez, and Susan~C Weber.
\newblock Stride--an integrated standards-based translational research
  informatics platform.
\newblock In \emph{AMIA Annual Symposium Proceedings}, volume 2009, page 391.
  American Medical Informatics Association, 2009.

\bibitem[Nguyen et~al.(2017)Nguyen, Tran, Wickramasinghe, and
  Venkatesh]{nguyen2017mathtt}
Phuoc Nguyen, Truyen Tran, Nilmini Wickramasinghe, and Svetha Venkatesh.
\newblock Deepr: A convolutional net for medical records.
\newblock \emph{IEEE journal of biomedical and health informatics}, 21\penalty0
  (1):\penalty0 22--30, 2017.

\bibitem[Rajkomar et~al.(2018)Rajkomar, Oren, Chen, Dai, Hajaj, Hardt, Liu,
  Liu, Marcus, Sun, et~al.]{GoogleEHR2018}
Alvin Rajkomar, Eyal Oren, Kai Chen, Andrew~M Dai, Nissan Hajaj, Michaela
  Hardt, Peter~J Liu, Xiaobing Liu, Jake Marcus, Mimi Sun, et~al.
\newblock Scalable and accurate deep learning with electronic health records.
\newblock \emph{npj Digital Medicine}, 1\penalty0 (1):\penalty0 18, 2018.

\bibitem[Suresh et~al.(2018)Suresh, Gong, and Guttag]{multisubgroups}
Harini Suresh, Jen~J Gong, and John Guttag.
\newblock Learning tasks for multitask learning: Heterogenous patient
  populations in the icu.
\newblock \emph{arXiv preprint arXiv:1806.02878}, 2018.

\bibitem[Wang et~al.(2014)Wang, Wang, and Hu]{multidisease}
Xiang Wang, Fei Wang, and Jianying Hu.
\newblock A multi-task learning framework for joint disease risk prediction and
  comorbidity discovery.
\newblock In \emph{Pattern Recognition (ICPR), 2014 22nd International
  Conference on}, pages 220--225. IEEE, 2014.

\bibitem[Yakovlev et~al.(2018)Yakovlev, Metsker, Kovalchuk, and
  Bologova]{acslos}
Alexey Yakovlev, Oleg Metsker, Sergey Kovalchuk, and Ekaterina Bologova.
\newblock Prediction of in-hospital mortality and length of stay in acute
  coronary syndrome patients using machine-learning methods.
\newblock \emph{Journal of the American College of Cardiology}, 71\penalty0
  (11):\penalty0 A242, 2018.

\end{thebibliography}

\end{document}